\documentclass{article}
\usepackage[preprint]{colm2026_conference}
\usepackage{microtype}
\usepackage{url}
\usepackage{booktabs}
\usepackage{tabularray}
\UseTblrLibrary{booktabs}
\usepackage{amsmath}
\usepackage{amssymb}
\usepackage{multirow}
\usepackage{array}
\usepackage{tabularx}
\usepackage{graphicx}
\usepackage[table]{xcolor}
\usepackage{pifont}
\definecolor{costL}{RGB}{232,242,252}  % lightest blue
\definecolor{costM}{RGB}{176,208,236}  % medium blue
\definecolor{costH}{RGB}{110,164,214}  % darkest blue
\newcommand{\cellH}{\cellcolor{costH}H}
\newcommand{\cellM}{\cellcolor{costM}M}
\newcommand{\cellL}{\cellcolor{costL}L}
\newcommand{\costL}{\colorbox{costL}{low-cost}}
\newcommand{\costM}{\colorbox{costM}{medium-cost}}
\newcommand{\costH}{\colorbox{costH}{high-cost}}
\usepackage{tikz}
\newcommand{\cmark}{\ding{51}}  % ✓
\newcommand{\xmark}{\ding{56}}  % ✗
\usetikzlibrary{arrows.meta,positioning,fit,shapes.geometric,calc}
\usepackage[edges]{forest}

\definecolor{hidden-draw}{RGB}{0,0,0}
\usepackage{lineno}
\usepackage{caption}
\usepackage{hyperref}       % hyperlinks
\definecolor{darkblue}{rgb}{0,0,0.5}
\definecolor{citecol}{RGB}{139,0,0}
\hypersetup{
	colorlinks=true,
	linkcolor=red,
	filecolor=blue,      
	urlcolor=magenta,
	citecolor={blue!50!green},
}

\newcolumntype{Y}{>{\raggedright\arraybackslash}X}

\title{Towards In-Parameter Memory Augmentation for Large Language Models}

\author{Haoyu Huang$^{\dagger}$\thanks{Contact: hhuangcp@connect.ust.hk},\quad Zhongwei Xie$^{\dagger}$,\quad Jiaxin Bai\setcounter{footnote}{2}\thanks{Corresponding Author, baijiaxin@hkbu.edu.hk} ,\quad Yisen Gao$^{\dagger}$,\quad Hong Ting Tsang$^{\dagger}$, \\\textbf{Wuganjing Song$^{\dagger}$,}\quad \textbf{Huihao Jing$^{\dagger}$,}\quad \textbf{Yufei Li$^{\dagger}$,}\quad \textbf{Yangqiu Song$^{\dagger}$} \\
$^{\dagger}$The Hong Kong University of Science and Technology, \\
$^{\ddagger}$Hong Kong Baptist University\\
Hong Kong SAR, China\\
\href{https://github.com/HKUST-KnowComp/Awesome-In-Parameter-Memory}
{
    \raisebox{-0.1ex}{\includegraphics[height=1.0em]{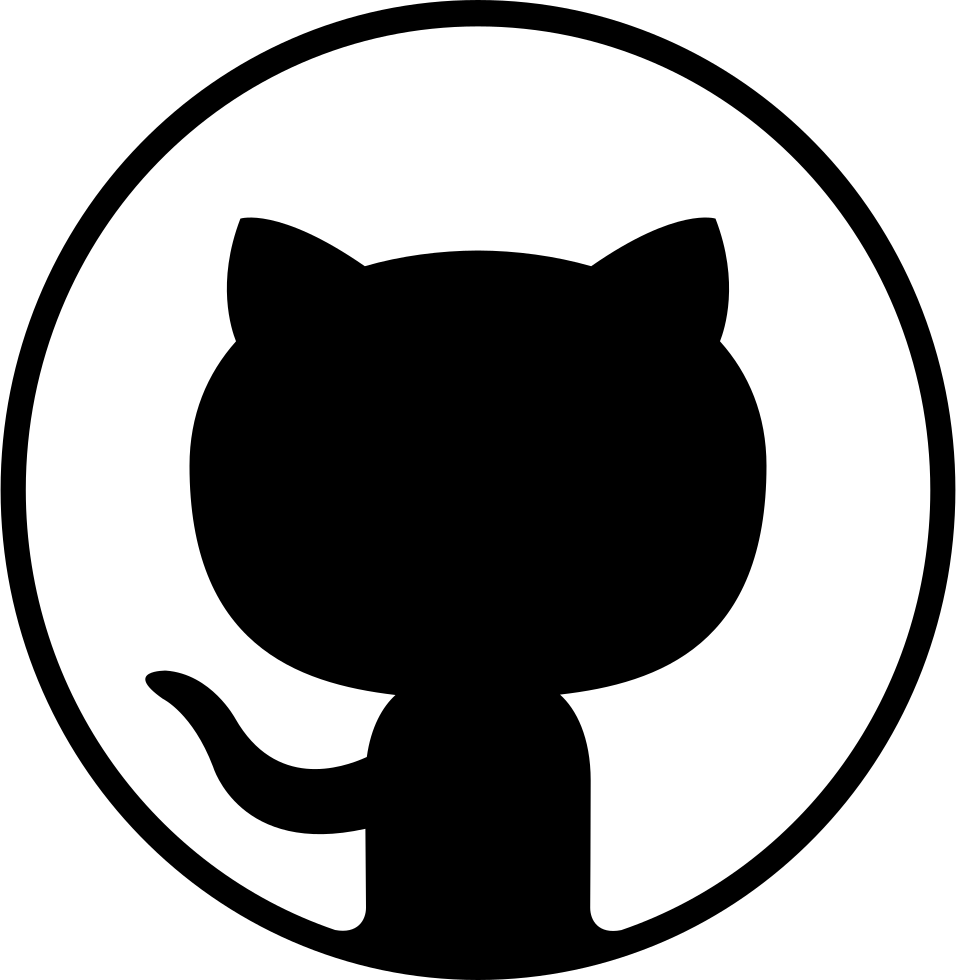}~Github Repository}
}
}

\begin{document}

\ifcolmsubmission
\linenumbers
\fi

\maketitle

\begin{abstract}
Recently Large Language Models (LLMs) and LLM-based agents increasingly need to incorporate knowledge acquired after pretraining, e.g., domain facts, user preferences, documents, and interaction experience. In-context learning (ICL) and ICL-based agent harness remain flexible, but they consume context capacity and incur repeated discretized encoding cost that grows with context length. \textbf{In-parameter memory} offers a complementary substrate: reusable memory information is represented in model parameters, adapters, or other parameter-like objects that are composed into the forward pass at inference time. This survey focuses on methods that augment LLMs with such parametric memory at deployment: a memory-bearing parameter object is plugged into the forward pass during inference, whether it is acquired before or during deployment. We organize the landscape with two orthogonal axes: \textbf{Parameter Placement}, which includes Embedding, Attention, FFN layers, or Hybrid when two or more layers are used; and \textbf{Parameter Acquisition Time}, which distinguishes methods whose memory object is acquired during deployment (online) from those acquired before it (offline). We clarify boundaries, conduct comparisons, and discuss open directions in interference, safety, co-design with ICL, and recursive self-improvement.
\end{abstract}

\begin{figure*}[h]
    \centering
    \includegraphics[width=0.95\linewidth]{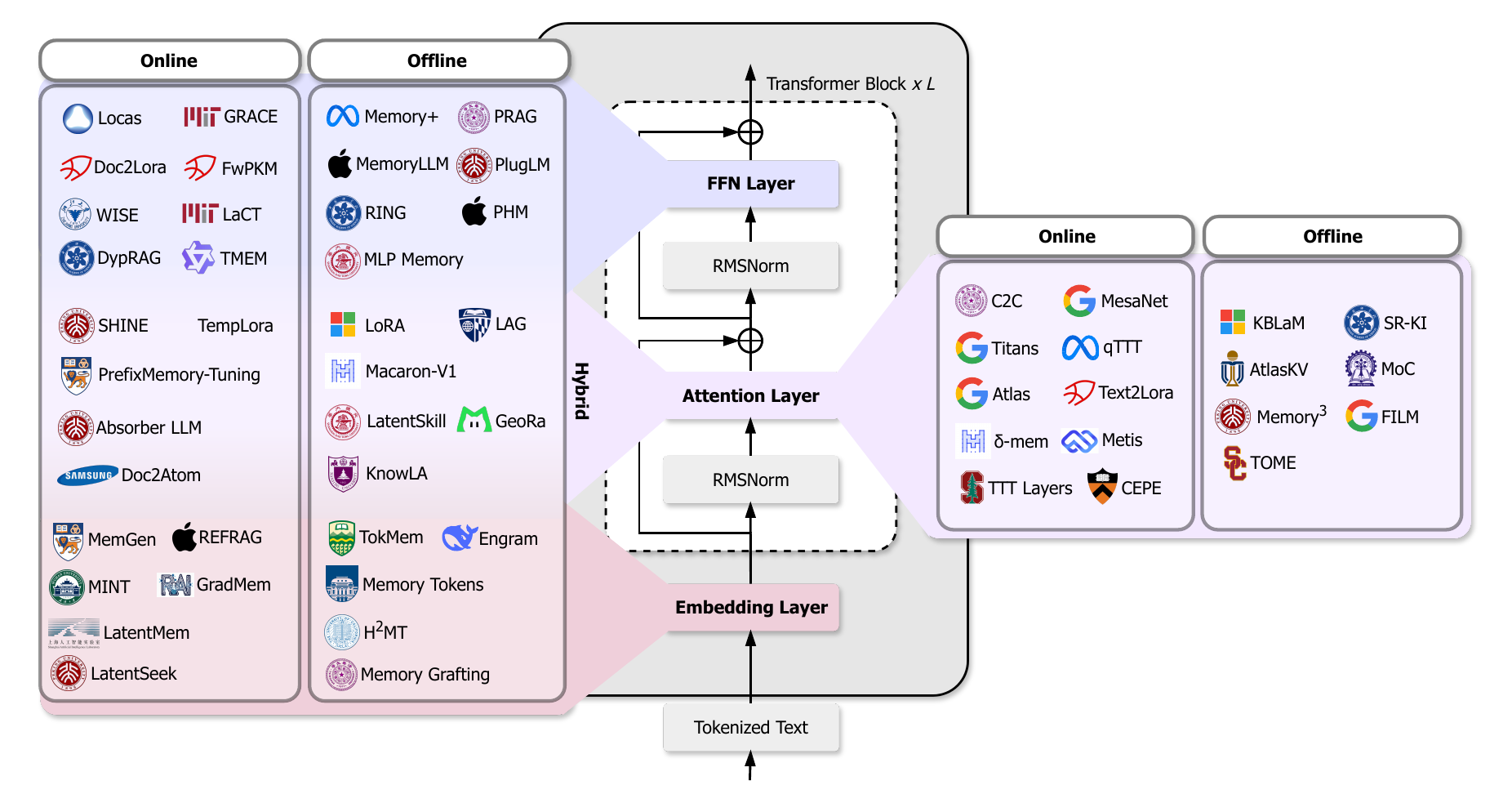}
    \vspace{-0.5cm}
    \caption{A demonstration of the representative methods in the two-dimensional taxonomy.}
    \label{fig:overview}
  \end{figure*}

  \section{Introduction}
  
  Large Language Models (LLMs) and LLM-based agents have demonstrated broad capabilities in natural language generation, reasoning, and tool use \citep{brown2020language,achiam2023gpt}. A persistent challenge is how to incorporate knowledge that is external to, or more recent than, pretraining, including new documents, domain-specific constraints, users' preferences, and experience accumulated over agentic interactions \citep{lewis2020retrieval,huang2026deeprefine,xu2026agent,han2026swe,ni2026trace2skill}, which we term as memory for LLMs in this survey.
  
  Currently most of the deployed systems expose this information as explicit textual tokens, which is also known as in-context learning (ICL) methods~\citep{dong2024survey}. ICL conditions a frozen model on textual demonstrations or retrieved documents as the memory, which is the most common paradigm used by various retrieval-augmented generation (RAG) methods~\citep{gao2023retrieval,fan2024survey,brown2020language,lewis2020retrieval,huang2025retrieval} and agents~\citep{ying2026security,deepseek-harness2026}. Though these ICL-based methods for memory augmentations are explainable, inspectable and easy to revise, their compute and latency grow with the number of retrieved tokens, and long contexts can exhaust a finite window~\citep{vaswani2017attention,liu2024lost}. Although sparse attention~\citep{zhang2025spargeattention,lai2026minimax,liu2025deepseek,yuan2025native,tang2024quest} and KV cache compression~\citep{rehg2024kv,kim2026kvzip,cai2024pyramidkv} alleviate but do not remove this scaling.
  
  In-parameter memory (IPM)~\citep{fleshman2025lora,huang2026atlaskv,lei2026delta,zhang2026metis} provides a complementary memory representation and augmentation paradigm. Instead of exposing all relevant information as explicit textual tokens, the system composes a compressed parametric memory object into the forward pass at inference time. The backbone parameters $\theta$ may stay frozen, or the augmentation step may write into them; what matters is that a memory object $\phi$ is acquired and then composed at deployment. The memory object may be acquired before deployment (offline) or produced and revised during deployment (online). Compared with ICL-based methods, this paradigm can avoid the quadratic cost of long-context prefilling~\citep{mcdermott2026lifelong} and reduce inference latency when memory is reused across queries. This motivates a significant shift from ICL-based to in-parameter memory augmentation~\citep{back2026understanding}.
  
  To highlight this ongoing shift, in this paper, we provide a focused overview and detailed discussion of the current research progress on in-parameter memory augmentation for LLMs.
  As shown in Figure~\ref{fig:overview}, we organize the in-parameter memory augmentation methods along two orthogonal dimensions:
  \textbf{(1) Parameter Placement} asks which part of the transformer block receives the parametric memory, including Embedding Layers, Attention Layers, FFN Layers, or Hybrid when two or more of these are used together.
  \textbf{(2) Parameter Acquisition Time} asks whether the deployed memory object is formed before the model starts serving (offline) or while it is serving (online).
  Concurrent architecture-centric surveys~\citep{zhoubian2026memory} cover memory broadly, including implicit computation-coupled memory (KV caches, SSMs, MoE) and parameter-efficient adaptation. We treat unmodified token-derived KV caches as token memory rather than in-parameter memory: they are a byproduct of prefilling, not an object produced by an independent acquisition operator $\mathcal{A}$. We index the remaining composable objects by target-module placement, acquisition timing, and deployment lifecycle (Section~\ref{sec:scope}).
  
  Our contributions are:
  \begin{itemize}
      \item We organize in-parameter memory (IPM) methods along two axes: parameter placement and acquisition time, and give a detailed landscape of the current research progress.
      \item We formulate IPM methods by a memory object $\phi$, an acquisition operator $\mathcal{A}$, and an augmentation operator $\mathcal{C}$ with explicit boundaries.
      \item We discuss open directions for IPM in interference, safety-aware acquisition, co-design with ICL, and recursive self-improvement.
  \end{itemize}
  
  \section{Formulation and Scope}
  \label{sec:scope}
  \subsection{Transformer Backbones and Memory-Able Sites}~\label{sec:transformer_backbone}
  Although nowadays there are many variants of the transformer backbone, their basic building blocks are still almost the same~\citep{xu2026deepseek, qiu2026design, team2026kimi}. The common transformer block composes an embedding layer, multi-head attention, a feed-forward network (FFN or FFN-based MoE, we use FFN to refer to both), residual connections, and normalization \citep{vaswani2017attention}. In-parameter memory does not aim to replace this anatomy, but injects a memory object $\phi$ into one or more of its sites. Two lines of evidence show where such injection is natural. First, FFN layers behave as key-value memories, and attribution studies link specific neurons and layers to factual associations~\citep{geva2021transformer,dai2022knowledge,meng2022locating}. Second, attention itself is an associative operator: ordinary self-attention accumulates token-derived key-value pairs in a growing cache, while linear-attention and fast-weight variants compress associations into a fixed-size state~\citep{schlag2021linear}. These observations motivate three placement families, defined by where IPM $\phi$ is stored or read:
  \begin{itemize}
      \item \textbf{Embedding layer:} $\phi$ enters as continuous prompt tokens, prefixes, or latent input states appended before the first block.
      \item \textbf{Attention layer:} $\phi$ participates in attention as keys/values, attention-side adapters, or matrix-valued associative states.
      \item \textbf{FFN layer:} $\phi$ is stored in feed-forward parameters or FFN-substituting projections.
  \end{itemize}
  A method is \textbf{Hybrid} when its deployed $\phi$ uses two or more of these families. The sites differ in read path, capacity, update cost, and compatibility with pretrained checkpoints, which is why placement is one important axis of the taxonomy.
  \begin{figure*}[!t]
    \centering
    \includegraphics[width=0.6\linewidth]{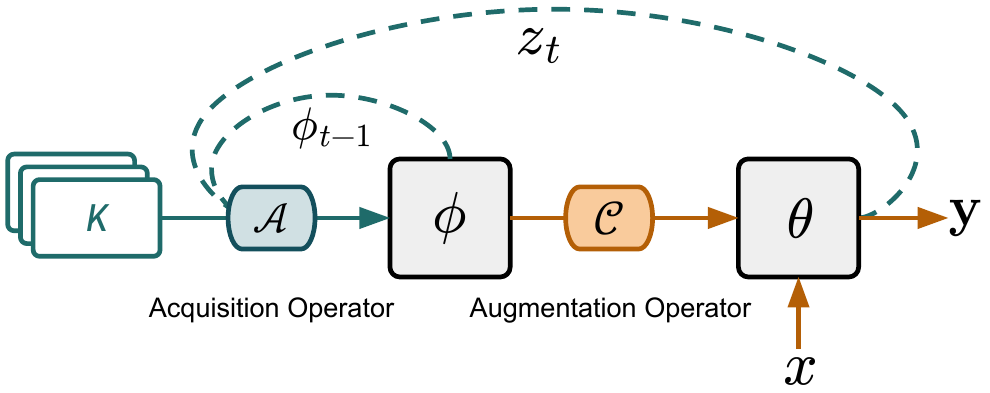}
    \caption{A demonstration of the operator formulation of in-parameter memory.}
    \label{fig:operator}
  \end{figure*}
  \label{sec:operator}
  \subsection{An Operator Formulation of In-Parameter Memory}
  Let $\theta$ denote the backbone parameters and $\phi$ a memory-bearing parameter object. As shown in Figure~\ref{fig:operator}, we characterize any in-parameter memory method by two operators: an \textbf{acquisition operator} $\mathcal{A}$ that produces $\phi$, and an \textbf{augmentation operator} $\mathcal{C}$ that composes $\phi$ into the forward pass. Acquisition and augmentation are distinct. For example, in LoRA generation methods~\citep{charakorn2025text,charakorn2026doc}, $\mathcal{A}$ is the hypernetwork that emits adapter matrices from a document or task description, while $\mathcal{C}$ is the inference-time rule $W\mapsto W{+}BA$ on the chosen modules. Placement in the taxonomy is the site of $\mathcal{C}$, not of $\mathcal{A}$.
  
  $\mathcal{A}$ has two forms. One-shot acquisition maps a source $K$ to IPM objects, $\phi=\mathcal{A}(K)$, which covers gradient optimization over continuous vectors~\citep{lester2021power,wang2026prefixmemory}, one-pass hypernetworks~\citep{charakorn2025text,charakorn2026doc}, and knowledge encoders that emit KV pairs~\citep{wang2025kblam,huang2026atlaskv}. Stateful acquisition rewrites the current IPM object from new evidence observed during deployment, $\phi_t=\mathcal{A}(\phi_{t-1},z_t)$, which covers recurrent write rules such as the delta-rule~\citep{lei2026delta} and the surprise-driven update~\citep{behrouz2026titans}. The $K$ in one-shot operator is usually static external memory and that in stateful operator is usually internal memory that are generated by the model itself. Both are acquisition: the second simply conditions on the previous $\phi$. The deployed model generates as
  \begin{equation}
  \phi_t=\mathcal{A}(\phi_{t-1},z_t),
  \end{equation}
  \begin{equation}
  y_t \sim p_{\mathcal{C}(\theta,\phi_t)}(y_t\mid x_{\leq t}),
  \end{equation}
  with $\phi_0=\mathcal{A}(K)$ for one-shot methods and $z_t$ the evidence observed at step $t$. $\mathcal{C}$ is the composition rule at the chosen site: augmenting $\phi$ to the frozen parameters $\theta$, adding it as a LoRA residual, reading it as extra keys/values in attention, adding it as virtual tokens~\citep{zhang2026memgen}, or directly updating latent representations~\citep{li2025seek}. For stateful methods, $\mathcal{C}$ runs on the current $\phi_t$. And how often $\mathcal{A}$ fires to update $\phi$ varies by method: per token~\citep{behrouz2026titans}, per segment~\citep{munkhdalai2024leave}, or on demand~\citep{kuratov2026gradmem,li2025seek}.
  
  The two operators play different roles in what follows. $\mathcal{A}$ and $\mathcal{C}$ first decide whether a method is in scope (Section~\ref{sec:scope_boundary}): $\phi$ must be parametric, produced by an independent $\mathcal{A}$, and augmented at deployment. Methods that pass are then located by where $\mathcal{C}$ reads $\phi$ and when $\mathcal{A}$ forms it (Section~\ref{sec:taxonomy}).

  \subsection{Scope and Relation to Neighboring Paradigms}
  \label{sec:scope_boundary}
  
  Table~\ref{tab:scope} applies the three tests to neighboring paradigms. Pluggable PEFT passes all three: the adapter is a continuous parameter object, trained by its own $\mathcal{A}$, and still plugged into the forward pass at deployment~\citep{hu2021lora,li2021prefix,lester2021power,liu2022ptuning}. Full-parameter SFT fails the last test, because the update is merged into $\theta$ before serving and leaves no distinct object for $\mathcal{C}$ to augment. An unmodified KV cache fails independent acquisition: its entries are a byproduct of prefilling discrete tokens, even though the stored tensors are continuous~\citep{rehg2024kv,kim2026kvzip,cai2024pyramidkv}. Encoded KV banks remain inside the boundary~\citep{huang2026atlaskv,wang2025kblam,yu2026sr,de2021mention}, because an encoder is an independent $\mathcal{A}$ and the LLM reads the resulting parameters rather than a raw token stream. A learned write rule that emits a parameter-like state is included on the same grounds~\citep{lei2026delta,feng2026inplace,von2026mesanet,bansal2026let}.
  
  \paragraph{Relation to continual learning.}
  Continual learning (CL) studies how to absorb new knowledge over a sequence of tasks while mitigating catastrophic forgetting~\citep{shi2025continual}. Its canonical mechanisms perform knowledge augmentation by offline training of the model parameters themselves, consolidating each new task before the model is served~\citep{kirkpatrick2017overcoming}. Recent LLM transfers converge toward IPM~\citep{lin2025continual,xu2026ring}, which maybe acquire $\phi$ in an offline stage but augment it at inference, so they enter our taxonomy as offline-acquired methods. The dividing line between CL and IPM is whether $\phi$ is still augmented at deployment.
  \begin{table*}[!t]
    \centering
    \small
  \caption{Scope relative to neighboring paradigms. \textbf{Parametric} asks whether the memory takes continuous parameter form. \textbf{Independent acquisition} asks whether it is produced by an independent acquisition operator $\mathcal{A}$ rather than as a byproduct of running the model. \textbf{Augmented at deployment} asks whether it enters the forward pass at deployment as a distinct object. A paradigm is included only when all three hold.}
  \label{tab:scope}
  \begin{booktabs}{
    colspec = {X[1.7,l] X[c] X[1.2,c] X[1.6,c] X[0.8,c]},
    width = \textwidth,
    cells = {valign=m},
    column{5} = {bg=gray!20},
    row{1} = {font=\bfseries},
    hline{1,Z} = {wd=0.08em},
    hline{2} = {wd=0.05em},
  }
    Paradigm & Parametric & Independent Acquisition & Augmented at Deployment & Included \\
    ICL-based RAG & \xmark & \xmark & \cmark & \xmark \\
    Pluggable PEFT & \cmark & \cmark & \cmark & \cmark \\
    Full Parameter SFT & \cmark & \cmark & \xmark & \xmark \\
    Unmodified KV Cache & \cmark & \xmark & \cmark & \xmark \\
  \end{booktabs}
\end{table*}
    \begin{table*}[!t]
      \centering
      \small
      \caption{The operational criteria and typical memory objects of different classes of methods in the two dimensions used to locate included methods.}
      \label{tab:definitions}
      \begin{booktabs}{
        colspec = {Q[l, wd=2cm] Q[l, wd=2.6cm] X[l, wd=3.5cm] X[l]},
        width = \textwidth,
        rowsep = 4pt,
        cells = {valign=m},
        cell{2}{1} = {r=2}{m},
        cell{4}{1} = {r=4}{m},
        row{1} = {font=\bfseries},
        hline{1,Z} = {wd=0.08em},
        hline{2,4} = {wd=0.05em},
      }
        Dimension & Class & Operational Criterion & Typical Memory Object \\
        \textbf{Parameter Acquisition Time} & Online & $\phi$ is generated or updated while the model is serving. & Fast Weights; Online Neural Memory; Adapters; Soft Tokens. \\
        & Offline & $\phi$ is formed before serving and held fixed, but still augmented at deployment. & Encoded KV Banks; Soft Prompts; Adapters; Memory Tables. \\
        \textbf{Parameter Placement} & Embedding Layer & $\phi$ is composed into the input embeddings. & Soft Prompts, Soft Tokens, Memory Tables. \\
        & Attention Layer & $\phi$ is composed inside attention. & Encoded KV Banks; Adapters; Fast Weights; Online Neural Memory; Soft Prompts. \\
        & FFN Layer & $\phi$ is composed inside the FFN. & Adapters; Memory Tables; Fast Weights; Online Neural Memory. \\
        & Hybrid & $\phi$ is composed at two or more of these sites. & Adapters, Online Neural Memory. \\
      \end{booktabs}
    \end{table*}
  \section{Two-Dimensional Taxonomy}
  \label{sec:taxonomy}
  
  To better understand and compare the properties of different IPM methods (Section~\ref{sec:compare}), we located the methods that pass Table~\ref{tab:scope} along two axes: where $\mathcal{C}$ augments $\phi$, and when $\mathcal{A}$ forms it, which form the \textbf{Parameter Placement} and \textbf{Parameter Acquisition Time} two taxonomy dimensions. Table~\ref{tab:definitions} states the operational criteria and typical memory objects in each dimension. Figure~\ref{fig:taxonomy} draws the taxonomy tree and representative methods. In this section, we will discuss how we define these two dimensions. The analysis about the methods in each class will be conducted in Section~\ref{sec:landscape}.
  
  \subsection{Dimension I: Parameter Placement}
  
  As shown in Table~\ref{tab:definitions}, the parameter placement dimension is defined by the site at which $\mathcal{C}$ reads $\phi$. And according to the families in Section~\ref{sec:transformer_backbone}, we can divide the parameter placement into four classes: embedding layers, attention layers, FFN layers, and Hybrid. And the typical memory objects for each class are also listed in Table~\ref{tab:definitions}. Different parameter placement corresponds to different memory objects. For example, the memory objects for the embedding layers are usually soft prompts~\citep{lester2021power,yi2025mint}, soft tokens~\citep{zhang2026memgen,wu2026tokmem}, and memory tables~\citep{cheng2026conditional,cheng2026memory}. And the memory objects for the attention layers are usually encoded KV banks~\citep{fu2026cache,huang2026atlaskv}, adapters~\citep{charakorn2025text}, fast weights~\citep{lei2026delta}, online neural memory~\citep{behrouz2026titans}, and soft prompts~\citep{wang2026prefixmemory}. And the memory objects for the FFN layers are usually adapters~\citep{su2025polyprag}, memory tables~\citep{xu2026ring}, online neural memory~\citep{zhang2026test} and fast weights~\citep{zhao2026fast}. And the memory objects for the Hybrid are usually adapters~\citep{mindlab2026macaronv1}, online neural memory~\citep{zhang2026absorber}.
  
  \subsection{Dimension II: Parameter Acquisition Time}
  
  The parameter acquisition time records when the deployed $\phi$ is formed: online or offline. The reference is whether $\phi$ is formed during or before the models' deployment. The offline-acquired IPM methods are usually based on one-shot acquisition operators, while the online-acquired IPM methods are usually based on stateful acquisition operators. Note that the acquisition time is not when the operator that produces it was trained. For example, a hypernetwork may be trained offline, yet the LoRA it emits for a deployment context is acquired online~\citep{tan2025dynamic}. The typical memory objects for each class are also listed in Table~\ref{tab:definitions}. Online objects are usually fast weights~\citep{lei2026delta}, online neural memory~\citep{behrouz2026titans}, adapters~\citep{charakorn2025text}, and soft tokens~\citep{zhang2026memgen}. Their augmentation operators are usually various forward processes. Offline objects are usually encoded KV banks~\citep{yu2026sr}, soft prompts~\citep{liu2022ptuning}, adapters~\citep{hu2021lora}, and memory tables~\citep{cheng2026conditional}. And their augmentation operators are usually their training processes.
  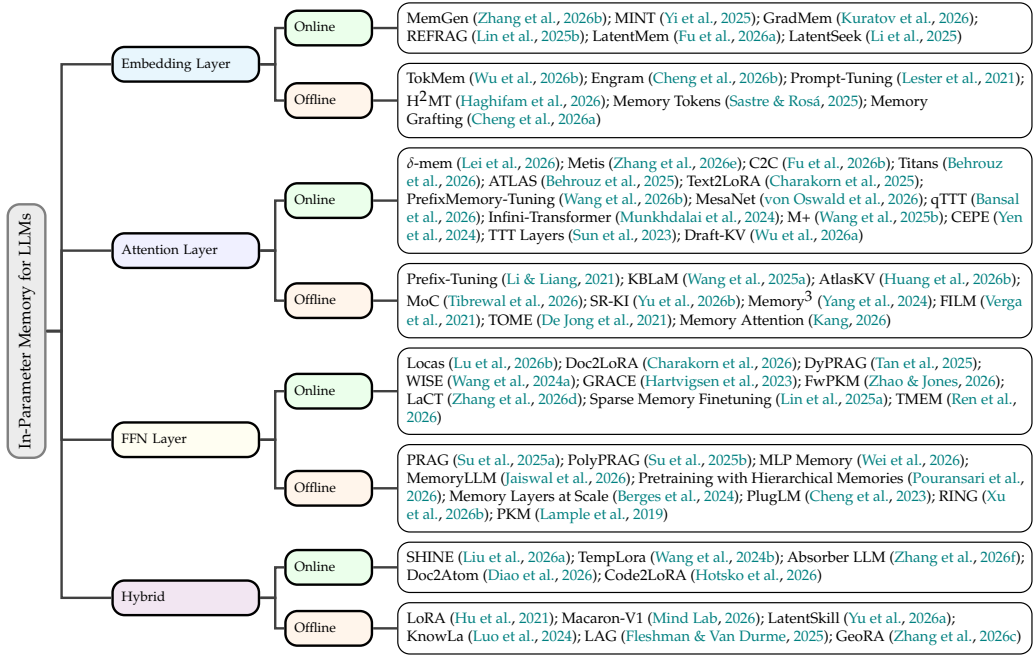
\begin{figure*}[t]
    \centering
    \resizebox{\textwidth}{!}{% Overall taxonomy: Level1 = Placement (4 layers); Level2 = Online/Offline; Level3 = methods.
% \parbox inside braces forces wrapping; forest text width can't reach through braces.
\tikzstyle{my-box}=[
    rectangle,
    rounded corners,
    text opacity=1,
    minimum height=0.1em,
    minimum width=0.1em,
    inner sep=2pt,
    align=left,
    fill opacity=.5,
]
\tikzstyle{leaf}=[my-box]
\newcommand{\leafbox}[1]{\parbox{7.2cm}{\fontsize{5}{6}\selectfont\raggedright #1}}
\begin{forest}
    forked edges,
    for tree={
        grow=east,
        reversed=true,
        anchor=west,
        parent anchor=east,
        child anchor=west,
        font={\fontsize{5.5}{6.5}\selectfont},
        rectangle,
        draw=hidden-draw,
        rounded corners,
        align=left,
        minimum width=0.1em,
        edge+={darkgray, line width=0.8pt},
        s sep=3pt,
        l sep=8pt,
        inner xsep=3pt,
        inner ysep=3pt,
        ver/.style={rotate=90, child anchor=north, parent anchor=south, anchor=center},
    },
    where level=1{text width=1.5cm,font={\fontsize{4.5}{5.5}\selectfont},align=center,tier=l1}{},
    where level=2{text width=0.8cm,font={\fontsize{4.5}{5.5}\selectfont},align=center,tier=l2}{},
    where level=3{
        tier=l3,
        edge path'={(!u.parent anchor) -- (.child anchor)},
    }{},
    [
    In-Parameter Memory for LLMs, draw=gray, color=gray!100, fill=gray!15, thick, text=black, ver,
    [Embedding Layer, black, fill=cyan!8, thick, text=black,
        [Online, color=black, fill=green!8, thick, text=black,
            [{\leafbox{MemGen~\citep{zhang2026memgen}; MINT~\citep{yi2025mint}; GradMem~\citep{kuratov2026gradmem}; REFRAG~\citep{lin2025refrag}; LatentMem~\citep{fu2026latentmem}; LatentSeek~\citep{li2025seek}}}]
        ]
        [Offline, color=black, fill=orange!8, thick, text=black,
            [{\leafbox{TokMem~\citep{wu2026tokmem}; Engram~\citep{cheng2026conditional}; Prompt-Tuning~\citep{lester2021power}; H$^2$MT~\citep{haghifam2026h}; Memory Tokens~\citep{sastre2025memory}; Memory Grafting~\citep{cheng2026memory}}}]
        ]
    ]
    [Attention Layer, black, fill=blue!6, thick, text=black,
        [Online, color=black, fill=green!8, thick, text=black,
            [{\leafbox{$\delta$-mem~\citep{lei2026delta}; Metis~\citep{zhang2026metis}; C2C~\citep{fu2026cache}; Titans~\citep{behrouz2026titans}; ATLAS~\citep{behrouz2025atlas}; Text2LoRA~\citep{charakorn2025text}; PrefixMemory-Tuning~\citep{wang2026prefixmemory}; MesaNet~\citep{von2026mesanet}; qTTT~\citep{bansal2026let}; Infini-Transformer~\citep{munkhdalai2024leave}; M+~\citep{wang2025m+}; CEPE~\citep{yen2024long}; TTT Layers~\citep{sun2023learning}; Draft-KV~\citep{wu2026draftkv}}}]
        ]
        [Offline, color=black, fill=orange!8, thick, text=black,
            [{\leafbox{Prefix-Tuning~\citep{li2021prefix}; KBLaM~\citep{wang2025kblam}; AtlasKV~\citep{huang2026atlaskv}; MoC~\citep{tibrewal2026mixture}; SR-KI~\citep{yu2026sr}; Memory$^{3}$~\citep{yang2024text}; FILM~\citep{verga-etal-2021-adaptable}; TOME~\citep{de2021mention}; Memory Attention~\citep{kang2026memory}}}]
        ]
    ]
    [FFN Layer, black, fill=yellow!6, thick, text=black,
        [Online, color=black, fill=green!8, thick, text=black,
            [{\leafbox{Locas~\citep{lu2026locas}; Doc2LoRA~\citep{charakorn2026doc}; DyPRAG~\citep{tan2025dynamic}; WISE~\citep{wang2024wise}; GRACE~\citep{hartvigsen2023aging}; FwPKM~\citep{zhao2026fast}; LaCT~\citep{zhang2026test}; Sparse Memory Finetuning~\citep{lin2025continual}; TMEM~\citep{ren2026scaling}}}]
        ]
        [Offline, color=black, fill=orange!8, thick, text=black,
            [{\leafbox{PRAG~\citep{su2025parametric}; PolyPRAG~\citep{su2025polyprag}; MLP Memory~\citep{wei2026mlp}; MemoryLLM~\citep{jaiswal2026memoryllm}; Pretraining with Hierarchical Memories~\citep{pouransari2026pretraining}; Memory Layers at Scale~\citep{berges2024memory}; PlugLM~\citep{cheng2023decouple}; RING~\citep{xu2026ring}; PKM~\citep{lample2019large}}}]
        ]
    ]
    [Hybrid, black, fill=violet!10, thick, text=black,
        [Online, color=black, fill=green!8, thick, text=black,
            [{\leafbox{SHINE~\citep{liu2026shine}; TempLora~\citep{wang2024greater}; Absorber LLM~\citep{zhang2026absorber}; Doc2Atom~\citep{diao2026doc}; Code2LoRA~\citep{hotsko2026code2lora}}}]
        ]
        [Offline, color=black, fill=orange!8, thick, text=black,
            [{\leafbox{LoRA~\citep{hu2021lora}; Macaron-V1~\citep{mindlab2026macaronv1}; LatentSkill~\citep{yu2026latentskill}; KnowLa~\citep{luo2024knowla}; LAG~\citep{fleshman2025lora}; GeoRA~\citep{zhang2026geora}}}]
        ]
    ]
    ]
\end{forest}}
    \caption{Overall taxonomy and representative methods of IPM for LLMs.}
    \label{fig:taxonomy}
    \end{figure*}
  \section{Method Landscape}
  \label{sec:landscape}
  
  Figure~\ref{fig:overview} gives an introductory demonstration and Figure~\ref{fig:taxonomy} lists the audited collection used throughout this section. We organize the narrative by placement (Level~1), then by acquisition time (Level~2), matching the taxonomy tree. In this section, we will discuss the details about the methods in each class.
  
  \subsection{Embedding Layer} 
  
  \paragraph{Online.}
  Most of the methods that augment online acquired IPM at embedding layers are based on soft tokens and soft prompts. MemGen~\citep{zhang2026memgen} equips LLMs with a weaver to synthesize the current hidden state into latent memory tokens that are prepended to the frozen backbone's input. GradMem~\citep{kuratov2026gradmem} compresses a context into a few prefix memory tokens via test-time gradient descent on a reconstruction loss, also with the backbone frozen. MINT~\citep{yi2025mint} maintains a bank of learnable soft prompts that is retrieved and updated online on test samples, adapting CLIP's input prompts while it serves. REFRAG~\citep{lin2025refrag} injects compressed chunk embeddings as input-side representations for efficient RAG decoding, selectively expanding chunks as needed. LatentMem~\citep{fu2026latentmem} gives each agent in a multi-agent system a memory composer that distills retrieved interaction trajectories into compact role-aware latent memory tokens. In all these cases the memory object is formed or updated during the inference time, so these methods are classified as online.
  
  \paragraph{Offline.}
  Prompt-tuning~\citep{lester2021power} optimizes continuous soft prompts before the deployment and prepend them at inference, so $\phi$ is composed offline. There are a line of methods using special tokens that are trained offline to integrate external memory. For example, TokMem~\citep{wu2026tokmem} trains procedural knowledge in specialized tokens offline and read at input. H$^2$MT~\citep{haghifam2026h} organizes hierarchical memory tree offline and routes each query top-down to read the relevant nodes. MemoryTokens~\citep{sastre2025memory} learns reversible sentence representations as a new token in the vocabulary. There are also a line of methods using memory tables or embedding tables that are trained offline for the models to look up. For example, Engram~\citep{cheng2026conditional} implements conditional $N$-gram lookup from input token identities. Memory Grafting~\citep{cheng2026memory} builds a frozen $N$-gram memory table from an offline grafting model and composes it at inference through longest-match lookup and gated injection. In these cases, $\phi$ is fixed during the episode but still enters the forward path at inference.
  
  \subsection{Attention Layer}
  
  \paragraph{Online.}
  There are a huge amount of test-time training (TTT) methods that augment online-acquired IPM at attention layers, which adapt online neural memory as their memory objects. For example, Titans~\citep{behrouz2026titans} and Atlas~\citep{behrouz2025atlas} learn attention-coupled neural memories at test time. MesaNet~\citep{von2026mesanet} and qTTT~\citep{bansal2026let} perform locally optimal or long-context TTT in attention-side state. Metis~\citep{zhang2026metis} pursues a memory-foundation model with online write or read. TTT Layers~\citep{sun2023learning} treat the hidden state as a model trained at test time with attention-side recurrence. There are also some works adapting fast weights as their memory objects. $\delta$-mem~\citep{lei2026delta} maintains a compact online associative state whose readout corrects attention of a frozen backbone. Infini-Transformer~\citep{munkhdalai2024leave} compresses past segments into a fixed-size associative memory matrix that each attention layer reads alongside the local cache. For KV banks as memory objects, these methods usually can obtain dense KV representations that are more than token-level memories. CEPE~\citep{yen2024long} encodes a long context chunk with a small encoder and lets the decoder read the encoded representations through added cross-attention layers. M+~\citep{wang2025m+} composes a query-conditioned memory bank at test time to extend the effective context. C2C~\citep{fu2026cache} treats the KV caches encoded by another LLM as the parametric memory and employs a learned cache fuser projects the sharer's KV cache into the receiver's attention. PrefixMemory-Tuning~\citep{wang2026prefixmemory} moves the prefix out of the attention head into an external query-conditioned module, fixing prefix-tuning's input-prefix tradeoff.
  
  \paragraph{Offline.}
  A typical offline-acquired IPM method at attention layers is prefix-tuning~\citep{li2021prefix}, for it learns continuous KV representations at each attention layer. There are also more flexible methods (e.g., KBLaM~\citep{wang2025kblam}, AtlasKV~\citep{huang2026atlaskv}, SR-KI~\citep{yu2026sr}) that learns cross-attention heads (projectors) as the acquisition operator $\mathcal{A}$ to project the representations of sentences encoded by a small freeze encoder to the LLMs' semantic space. They can obtain KV representations of new sentences on the fly. But their knowledge is organized offline, so they are classified as offline methods. MoC~\citep{tibrewal2026mixture} learns chapter-level memory modules offline that are attached at inference. Memory$^{3}$~\citep{yang2024text} converts a knowledge base into offline trained sparse attention key-value memories and retrieves them into self-attention at inference. Memory Attention~\citep{kang2026memory} drops the value projection and builds each attention value as the contextual key plus a lookup from a layer-specific token embedding table.
  
  \subsection{FFN Layer}
  
  \paragraph{Online.}
  The methods belonging to this class are usually based on hypernetworks to generate MLP LoRA adapters on the fly. Doc2LoRA~\citep{charakorn2026doc} maps a document to LoRA adapters in one pass. The reported implementation adapts MLP down-projections and is therefore FFN. DyPRAG~\citep{tan2025dynamic} dynamically generates parametric adapters from a retrieved document at test time for knowledge enhancement, where $\phi$ is materialized during the deployment. Locas~\citep{lu2026locas} and WISE~\citep{wang2024wise} provide locally supported or side-memory mechanisms for continual FFN-side updates at deployment. GRACE~\citep{hartvigsen2023aging} wraps an FFN projection with a codebook of key-value adaptors, applying a stored value only when the layer activation falls near a cached key. FwPKM~\citep{zhao2026fast} introduces fast-weight product-key memory on the feed-forward path. LaCT~\citep{zhang2026test} revisits test-time training with FFN-side state updates.
  
  \paragraph{Offline.}
  For offline methods, they usually synthesize some QA pairs from some documents or knowledge bases and train LoRA adapters on the FFN layers with them. For example, PRAG and PolyPRAG~\citep{su2025parametric,su2025polyprag} encode retrieved knowledge into FFN-side parameters offline and then plug them in at inference. Other methods like MLP Memory~\citep{wei2026mlp} pretrain an MLP at FFN layers to imitate a $k$-NN retriever's behavior, which internalize retrieval patterns without explicit document access. Similarly, RING~\citep{xu2026ring} injects new corpora into mixture-of-memory experts and learns, via reinforcement, the routing-and-search policy that reads them at inference. MemoryLLM~\citep{wei2026mlp,jaiswal2026memoryllm} trains the FFN only on context-free token embeddings, so the whole layer collapses into a precomputable token-wise lookup. Sparse memory finetuning~\citep{berges2024memory} writes new facts into the most selectively activated slots of a memory-layer model via masked gradient updates, leaving the addressable table to be read by top-$k$ lookup at inference. These compose $\phi$ at deployment and are offline-acquired core.
  
  \subsection{Hybrid}
  Unlike the FFN-side methods above, Hybrid methods train linear layers at more than one site: not only inside the FFN, but also at other modules such as attention heads.
  \paragraph{Online.}
  Temp-LoRA~\citep{wang2024greater} trains temporary multi-site LoRA during inference. SHINE~\citep{liu2026shine} maps a context to per-layer LoRA adapters in one forward pass of an in-context hypernetwork, so later queries answer from those adapters without rereading the context. AbsorberLLM~\citep{zhang2026absorber} synchronizes causal state for test-time training across sites. Doc2Atom~\citep{diao2026doc} compiles documents into composable semantically typed memory atoms spanning multiple modules.
  
  \paragraph{Offline.}
  Standard LoRA~\citep{hu2021lora} is hybrid when attention and FFN maps are adapted together. Macaron-V1~\citep{mindlab2026macaronv1} routes each turn to one specialist LoRA on a frozen base, so skills stay plug-in adapters. KnowLa~\citep{luo2024knowla} injects frozen knowledge-graph entity embeddings into hidden states through a small adapter trained alongside LoRA. LatentSkill~\citep{yu2026latentskill} compiles textual skills into LoRA adapters offline that are mounted on a frozen backbone at inference.
  
  \section{Property-Based Comparison}
  \label{sec:compare}
  In this section, we will compare the methods in each class based on their properties, for example, task purpose, materialization, persistence, and memory write or read cost, which decide how such an IPM object can be used. And we also list these properties in Table~\ref{tab:props} for each representative method. We also summarize each representative method's parameter placement of $\mathcal{C}$, the memory object, and the acquisition operator $\mathcal{A}$ in Table~\ref{tab:comparison}. Note that we use qualitative bands because the underlying papers rarely report comparable FLOPs or wall-clock numbers.
  
  As the tables suggest, online rows are almost always per-request or per-token and ephemeral or session-scoped, where generated adapters and soft prompts pay a \costM{} one-pass write for a fresh object, while recurrent fast weights and online neural memory pay a \costL{} per-token write to keep a mutable state for long-context modeling. Offline rows are almost always pre-deployment and persistent, where they pay a \costH{} one-time write to freeze knowledge into soft prompts, KV banks, or memory tables, and then read at \costL{} cost. Methods with the same IPM object type also behaves differently by parameter acquisition time. Adapters are cheap when generated online, but expensive when trained offline. Encoded KV banks are composed per request in CEPE and M+, but precomputed in KBLaM and AtlasKV.
  
  \subsection{Trade-offs}
  
  Based on the above analysis, we can summarize the trade-offs between IPM methods. For the parameter acquisition time axis, online acquisition adapts immediately but pays higher write-time compute. Offline acquisition is more stable and validated, yet cannot absorb a new stream without an offline training process. For the parameter placement axis, embedding-side objects are the cheapest to compose, because they enter through the input stream with the backbone frozen, but their influence is indirect, mediated by every layer above. Attention-side objects are content-addressable: the query-key reads support both associative lookup over large banks and recurrently updated states, at the cost of coupling to head structure and attention budget. FFN-side objects write into the layers where factual associations are stored, so edits can directly overwrite specific behaviors. However, the read is position-wise, which means each token consults the memory independently, with no content-based selection among entries and sequential writes to shared weights interfere. Hybrid stacks compose objects across sites and are the most expressive, shaping both what the model reads and how it computes. However, observed behaviors can not be traced to a single component, so updates must be coordinated across heterogeneous objects.
  
  Besides, ICL and ICL-based RAG remain preferable for the tasks needing exact quotation, provenance, targeted deletion, or human revision, which are trivial for a text store, hard once text is compressed and abstracted into $\phi$.
  
  \section{Open Directions}
  \label{sec:open}
  
  The comparison above exposes some gaps in the current landscape. We highlight four open directions for the IPM paradigm.
  
  \textbf{Interference and forgetting.}
  Sequential edits and recurrent delta-rule writes overwrite earlier associations, and the damage grows with the update sequence: edited weights increase in norm and disrupt general-ability directions, while a fixed state interferes along the current key~\citep{gupta2025lifelong,lyu2026evoedit,zhang2026spectral,lei2026delta,hatamizadeh2026gated2}. Subspace projections, norm constraints, and erase or write gates slow this collapse, but they keep a growing record of past keys or gate only the current key, so conflict detection and fixed-capacity allocation over a long stream remain open~\citep{wang2026labyrinth}.
  
  \textbf{Safety-aware acquisition.}
  One write from untrusted context is later retrieved as trusted knowledge, and current memory systems do not record provenance~\citep{dash2026untrusted,zou2026poison}. When the write is parametric, IPM objects can implant the behavior in weights, and behavioral unlearning leaves traces, so update budgets, provenance-conditioned write gates, and privacy-aware consolidation remain open~\citep{chen2026lora,hong2025intrinsic,wang2026agentic}.
  
  \textbf{Co-design with ICL.}
  ICL keeps memory as textual tokens, which can represent more detailed information. Although compressed memory objects in IPM can lead to efficient augmentation, there are still cases that need fine-grained memory access. So the open direction is a co-design in which $\mathcal{A}$ decides what stays in context and what enters $\phi$, and $\mathcal{C}$ reads both in one forward pass.
  
  \textbf{Model-native recursive self-improvement (RSI).}
  Most recursive self-improvement today lives in the agent harness, where prompts~\citep{fernando2023promptbreeder}, skills~\citep{liu2026skillrevise}, knowledge bases~\citep{tang2025agent}, and tools~\citep{lu2026beyond} are revised outside the model~\citep{ni2026trace2skill,deepseek-harness2026}. A model-native version would let $\mathcal{A}$ write useful deployment experience into $\phi$ itself, so the backbone model improves in-parameter and the gap between offline training and online inference narrows.
  
  \section{Conclusion}
  
  This survey analyzes in-parameter memory augmentation for LLMs, which is a more efficient and model-native way to augment LLMs than the ICL paradigm. We organize in-parameter memory along parameter placement and acquisition time, and lay out the resulting landscape. We formulate the paradigm with a memory object, an acquisition operator, and a composition operator, and use them to draw explicit boundaries. We also discuss the open directions for the IPM paradigm to help future parametric memory systems.
  
  \section{Limitations}
  
  This survey has the following limitations. First, it only considers the IPM paradigm under text-modality. There are also plenty of works applying the IPM paradigm to other modalities, such as vision, audio, and video. Second, it does not include IPM designs built into newer LLM architectures, such as Mixture of Latent Experts (MLE), Multi-head Latent Attention (MLA), and Manifold-Constrained Hyper-Connections (mHC). Third, the comparison does not under numerical measurements. Write cost, read cost, and retention are qualitative bands inferred from mechanisms, because the source papers rarely share a benchmark or comparable measurements.
  
\bibliographystyle{colm2026_conference}
\bibliography{colm2026_conference}

\appendix

\section{Comparison Tables}
\label{sec:appendix-tables}

Table~\ref{tab:comparison} and Table~\ref{tab:props} are given in this appendix. Table~\ref{tab:comparison} records, for every method in Figure~\ref{fig:taxonomy}, where $\phi$ is placed, what memory object it is, and which acquisition operator $\mathcal{A}$ produces it. Table~\ref{tab:props} records how each of those objects is used at deployment: the task, whether it persists, and the qualitative write and read cost. Section~\ref{sec:compare} discusses both tables.

\begin{table}[t]
  \centering
  \scriptsize
  \setlength{\tabcolsep}{3.2pt}
  \caption{Memory object and acquisition operator of every method in Figure~\ref{fig:taxonomy}. Memory objects use the seven types of Table~\ref{tab:definitions}. All rows are core; permanently merged comparators are excluded.}
  \label{tab:comparison}
  \begin{tabularx}{\textwidth}{@{}
    >{\setlength{\hsize}{1.50\hsize}\raggedright\arraybackslash}X
    >{\setlength{\hsize}{0.42\hsize}\raggedright\arraybackslash}X
    >{\setlength{\hsize}{0.78\hsize}\raggedright\arraybackslash}X
    >{\setlength{\hsize}{1.2\hsize}\raggedright\arraybackslash}X
  @{}}
  \toprule
  \textbf{Method} & \textbf{Placement} & \textbf{Memory object} & \textbf{Acquisition operator $\mathcal{A}$} \\
  \midrule
  \multicolumn{4}{@{}l}{\textbf{Embedding --- Online}} \\
  \addlinespace[1pt]
  MemGen \citep{zhang2026memgen} & Embedding & Soft Tokens & RL-trained trigger + LoRA weaver \\
  MINT \citep{yi2025mint} & Embedding & Soft Prompts & Test-time bank update \\
  GradMem \citep{kuratov2026gradmem} & Embedding & Soft Tokens & Test-time gradient descent \\
  REFRAG \citep{lin2025refrag} & Embedding & Soft Tokens & Selective expansion policy \\
  LatentMem \citep{fu2026latentmem} & Embedding & Soft Tokens & LMPO-trained composer \\
  LatentSeek \citep{li2025seek} & Embedding & Soft Tokens & Test-time policy gradient \\
  \addlinespace[2pt]
  \multicolumn{4}{@{}l}{\textbf{Embedding --- Offline}} \\
  \addlinespace[1pt]
  TokMem \citep{wu2026tokmem} & Embedding & Soft Tokens & Offline training \\
  Engram \citep{cheng2026conditional} & Embedding & Memory Tables & Offline training \\
  Prompt-Tuning \citep{lester2021power} & Embedding & Soft Prompts & Gradient optimization \\
  H$^2$MT \citep{haghifam2026h} & Embedding & Memory Tables & Bottom-up aggregation \\
  Memory Tokens \citep{sastre2025memory} & Embedding & Soft Tokens & Gradient optimization \\
  Memory Grafting \citep{cheng2026memory} & Embedding & Memory Tables & Offline grafting \\
  \midrule
  \multicolumn{4}{@{}l}{\textbf{Attention --- Online}} \\
  \addlinespace[1pt]
  $\delta$-mem \citep{lei2026delta} & Attention & Fast Weights & Gated delta rule \\
  Metis \citep{zhang2026metis} & Attention & Online Neural Memory & Online write/read \\
  C2C \citep{fu2026cache} & Attention & Encoded KV Banks & Cache-to-cache transfer \\
  Titans \citep{behrouz2026titans} & Attention & Online Neural Memory & Surprise-driven update \\
  ATLAS \citep{behrouz2025atlas} & Attention & Online Neural Memory & Context-aware update \\
  Text2LoRA \citep{charakorn2025text} & Attention & Adapters & One-pass hypernetwork \\
  PrefixMemory-Tuning \citep{wang2026prefixmemory} & Attention & Soft Prompts & Test-time prefix generation \\
  MesaNet \citep{von2026mesanet} & Attention & Online Neural Memory & Test-time training \\
  qTTT \citep{bansal2026let} & Attention & Online Neural Memory & Test-time training \\
  Infini-Transformer \citep{munkhdalai2024leave} & Attention & Fast Weights & Compressive memory update \\
  M+ \citep{wang2025m+} & Attention & Encoded KV Banks & Test-time memory composition \\
  CEPE \citep{yen2024long} & Attention & Encoded KV Banks & Context encoder \\
  TTT Layers \citep{sun2023learning} & Attention & Online Neural Memory & Test-time training \\
  Draft-KV \citep{wu2026draftkv} & Attention & Encoded KV Banks & Draft-then-verify \\
  \addlinespace[2pt]
  \multicolumn{4}{@{}l}{\textbf{Attention --- Offline}} \\
  \addlinespace[1pt]
  Prefix-Tuning \citep{li2021prefix} & Attention & Soft Prompts & Gradient optimization \\
  KBLaM \citep{wang2025kblam} & Attention & Encoded KV Banks & Knowledge encoder \\
  AtlasKV \citep{huang2026atlaskv} & Attention & Encoded KV Banks & Knowledge encoder \\
  MoC \citep{tibrewal2026mixture} & Attention & Memory Tables & Offline training \\
  SR-KI \citep{yu2026sr} & Attention & Encoded KV Banks & Offline training \\
  Memory$^{3}$ \citep{yang2024text} & Attention & Encoded KV Banks & Sparse KV selection \\
  FILM \citep{verga-etal-2021-adaptable} & Attention & Encoded KV Banks & Offline training \\
  TOME \citep{de2021mention} & Attention & Encoded KV Banks & Offline training \\
  Memory Attention \citep{kang2026memory} & Attention & Memory Tables & Offline training \\
  \midrule
  \multicolumn{4}{@{}l}{\textbf{FFN --- Online}} \\
  \addlinespace[1pt]
  Locas \citep{lu2026locas} & FFN & Adapters & Continual online edit \\
  Doc2LoRA \citep{charakorn2026doc} & FFN & Adapters & One-pass hypernetwork \\
  DyPRAG \citep{tan2025dynamic} & FFN & Adapters & Test-time parametric generation \\
  WISE \citep{wang2024wise} & FFN & Adapters & Side-memory edit \\
  GRACE \citep{hartvigsen2023aging} & FFN & Adapters & Codebook update \\
  FwPKM \citep{zhao2026fast} & FFN & Fast Weights & Outer-product write \\
  LaCT \citep{zhang2026test} & FFN & Online Neural Memory & Test-time training \\
  Sparse Memory Finetuning \citep{lin2025continual} & FFN & Memory Tables & Selective slot update \\
  TMEM \citep{ren2026scaling} & FFN & Memory Tables & Test-time memory write \\
  \addlinespace[2pt]
  \multicolumn{4}{@{}l}{\textbf{FFN --- Offline}} \\
  \addlinespace[1pt]
  PRAG \citep{su2025parametric} & FFN & Adapters & Offline parametric RAG \\
  PolyPRAG \citep{su2025polyprag} & FFN & Adapters & Offline parametric RAG \\
  MLP Memory \citep{wei2026mlp} & FFN & Adapters & Offline training \\
  MemoryLLM \citep{jaiswal2026memoryllm} & FFN & Memory Tables & Context-free FFN training \\
  Hierarchical Memories \citep{pouransari2026pretraining} & FFN & Memory Tables & Offline pretraining \\
  Memory Layers at Scale \citep{berges2024memory} & FFN & Memory Tables & Offline training \\
  PlugLM \citep{cheng2023decouple} & FFN & Adapters & Offline training \\
  RING \citep{xu2026ring} & FFN & Memory Tables & RL routing policy \\
  PKM \citep{lample2019large} & FFN & Memory Tables & Offline training \\
  \midrule
  \multicolumn{4}{@{}l}{\textbf{Hybrid --- Online}} \\
  \addlinespace[1pt]
  SHINE \citep{liu2026shine} & Hybrid & Adapters & In-context hypernetwork \\
  Temp-LoRA \citep{wang2024greater} & Hybrid & Adapters & Inference-time training \\
  Absorber LLM \citep{zhang2026absorber} & Hybrid & Online Neural Memory & Test-time sync \\
  Doc2Atom \citep{diao2026doc} & Hybrid & Adapters & Document compiler \\
  Code2LoRA \citep{hotsko2026code2lora} & Hybrid & Adapters & Hypernetwork \\
  \addlinespace[2pt]
  \multicolumn{4}{@{}l}{\textbf{Hybrid --- Offline}} \\
  \addlinespace[1pt]
  LoRA \citep{hu2021lora} & Hybrid & Adapters & Gradient optimization \\
  Macaron-V1 \citep{mindlab2026macaronv1} & Hybrid & Adapters & Offline training \\
  LatentSkill \citep{yu2026latentskill} & Hybrid & Adapters & Offline compilation \\
  KnowLa \citep{luo2024knowla} & Hybrid & Adapters & Gradient optimization \\
  LAG \citep{fleshman2025lora} & Hybrid & Adapters & Gradient optimization \\
  GeoRA \citep{zhang2026geora} & Hybrid & Adapters & Gradient optimization \\
  \bottomrule
  \end{tabularx}
  \end{table}

    \begin{table}[t]
    \centering
    \scriptsize
    \setlength{\tabcolsep}{3.2pt}
    \caption{Deployment properties of the methods in Table~\ref{tab:comparison}. Write/read cost are qualitative bands (L/M/H) inferred from the mechanism.}
    \label{tab:props}
    \begin{tabularx}{\textwidth}{@{}
        >{\setlength{\hsize}{2.40\hsize}\raggedright\arraybackslash}X
        >{\setlength{\hsize}{1.80\hsize}\raggedright\arraybackslash}X
        >{\centering\arraybackslash\setlength{\hsize}{0.60\hsize}}X
        >{\centering\arraybackslash\setlength{\hsize}{0.40\hsize}}X
        >{\centering\arraybackslash\setlength{\hsize}{0.40\hsize}}X
        >{\centering\arraybackslash\setlength{\hsize}{0.40\hsize}}X
    @{}}
    \toprule
    \textbf{Method} & \textbf{Task purpose} & \textbf{Material.} & \textbf{Persist.} & \textbf{Write} & \textbf{Read} \\
    \midrule
    \multicolumn{6}{@{}l}{\textbf{Embedding --- Online}} \\
    \addlinespace[1pt]
    MemGen \citep{zhang2026memgen} & Agent self-evolution & Per-request & Ephem. & \cellM & \cellL \\
    MINT \citep{yi2025mint} & Task adaptation & Per-request & Ephem. & \cellM & \cellL \\
    GradMem \citep{kuratov2026gradmem} & Long-context modeling & Per-request & Ephem. & \cellM & \cellL \\
    REFRAG \citep{lin2025refrag} & Knowledge injection & Per-request & Ephem. & \cellM & \cellL \\
    LatentMem \citep{fu2026latentmem} & Multi-agent Communication & Per-request & Ephem. & \cellM & \cellL \\
    LatentSeek \citep{li2025seek} & Reasoning & Per-request & Ephem. & \cellM & \cellL \\
    \addlinespace[2pt]
    \multicolumn{6}{@{}l}{\textbf{Embedding --- Offline}} \\
    \addlinespace[1pt]
    TokMem \citep{wu2026tokmem} & Knowledge injection & Pre-deploy & Persist. & \cellH & \cellL \\
    Engram \citep{cheng2026conditional} & Language modeling & Pre-deploy & Persist. & \cellH & \cellL \\
    Prompt-Tuning \citep{lester2021power} & Task adaptation & Pre-deploy & Persist. & \cellH & \cellL \\
    H$^2$MT \citep{haghifam2026h} & Language modeling & Pre-deploy & Persist. & \cellH & \cellL \\
    Memory Tokens \citep{sastre2025memory} & Sequence reconstruction & Pre-deploy & Persist. & \cellH & \cellL \\
    Memory Grafting \citep{cheng2026memory} & Language modeling & Pre-deploy & Persist. & \cellH & \cellL \\
    \midrule
    \multicolumn{6}{@{}l}{\textbf{Attention --- Online}} \\
    \addlinespace[1pt]
    $\delta$-mem \citep{lei2026delta} & Long-context modeling & Per-token & Session & \cellL & \cellL \\
    Metis \citep{zhang2026metis} & Continual learning & Per-token & Session & \cellL & \cellL \\
    C2C \citep{fu2026cache} & Multi-agent Communication & Per-request & Ephem. & \cellM & \cellL \\
    Titans \citep{behrouz2026titans} & Long-context modeling & Per-token & Session & \cellL & \cellL \\
    ATLAS \citep{behrouz2025atlas} & Long-context modeling & Per-token & Session & \cellL & \cellL \\
    Text2LoRA \citep{charakorn2025text} & Task adaptation & Per-request & Ephem. & \cellM & \cellL \\
    PrefixMemory-Tuning \citep{wang2026prefixmemory} & Task adaptation & Per-request & Ephem. & \cellM & \cellL \\
    MesaNet \citep{von2026mesanet} & Long-context modeling & Per-token & Session & \cellL & \cellL \\
    qTTT \citep{bansal2026let} & Long-context modeling & Per-token & Session & \cellL & \cellL \\
    Infini-Transformer \citep{munkhdalai2024leave} & Long-context modeling & Per-token & Session & \cellL & \cellL \\
    M+ \citep{wang2025m+} & Long-context modeling & Per-request & Ephem. & \cellM & \cellL \\
    CEPE \citep{yen2024long} & Long-context modeling & Per-request & Ephem. & \cellM & \cellL \\
    TTT Layers \citep{sun2023learning} & Long-context modeling & Per-token & Session & \cellL & \cellL \\
    Draft-KV \citep{wu2026draftkv} & Multi-agent Communication & Per-request & Ephem. & \cellM & \cellL \\
    \addlinespace[2pt]
    \multicolumn{6}{@{}l}{\textbf{Attention --- Offline}} \\
    \addlinespace[1pt]
    Prefix-Tuning \citep{li2021prefix} & Task adaptation & Pre-deploy & Persist. & \cellH & \cellL \\
    KBLaM \citep{wang2025kblam} & Knowledge injection & Pre-deploy & Persist. & \cellH & \cellL \\
    AtlasKV \citep{huang2026atlaskv} & Knowledge injection & Pre-deploy & Persist. & \cellH & \cellL \\
    MoC \citep{tibrewal2026mixture} & Knowledge injection & Pre-deploy & Persist. & \cellH & \cellL \\
    SR-KI \citep{yu2026sr} & Knowledge injection & Pre-deploy & Persist. & \cellH & \cellL \\
    Memory$^{3}$ \citep{yang2024text} & Knowledge injection & Pre-deploy & Persist. & \cellH & \cellL \\
    FILM \citep{verga-etal-2021-adaptable} & Knowledge injection & Pre-deploy & Persist. & \cellH & \cellL \\
    TOME \citep{de2021mention} & Knowledge injection & Pre-deploy & Persist. & \cellH & \cellL \\
    Memory Attention \citep{kang2026memory} & Language modeling & Pre-deploy & Persist. & \cellH & \cellL \\
    \midrule
    \multicolumn{6}{@{}l}{\textbf{FFN --- Online}} \\
    \addlinespace[1pt]
    Locas \citep{lu2026locas} & Lifelong editing & Per-request & Ephem. & \cellM & \cellL \\
    Doc2LoRA \citep{charakorn2026doc} & Document QA & Per-request & Ephem. & \cellM & \cellL \\
    DyPRAG \citep{tan2025dynamic} & Document QA & Per-request & Ephem. & \cellM & \cellL \\
    WISE \citep{wang2024wise} & Lifelong editing & Per-request & Ephem. & \cellM & \cellL \\
    GRACE \citep{hartvigsen2023aging} & Lifelong editing & Per-request & Ephem. & \cellM & \cellL \\
    FwPKM \citep{zhao2026fast} & Language modeling & Per-token & Session & \cellL & \cellL \\
    LaCT \citep{zhang2026test} & Long-context modeling & Per-token & Session & \cellL & \cellL \\
    Sparse Memory Finetuning \citep{lin2025continual} & Continual learning & Per-request & Ephem. & \cellM & \cellL \\
    TMEM \citep{ren2026scaling} & Agent self-evolution & Per-request & Ephem. & \cellM & \cellL \\
    \addlinespace[2pt]
    \multicolumn{6}{@{}l}{\textbf{FFN --- Offline}} \\
    \addlinespace[1pt]
    PRAG \citep{su2025parametric} & Document QA & Pre-deploy & Persist. & \cellH & \cellL \\
    PolyPRAG \citep{su2025polyprag} & Document QA & Pre-deploy & Persist. & \cellH & \cellL \\
    MLP Memory \citep{wei2026mlp} & Knowledge injection & Pre-deploy & Persist. & \cellH & \cellL \\
    MemoryLLM \citep{jaiswal2026memoryllm} & Language modeling & Pre-deploy & Persist. & \cellH & \cellL \\
    Hierarchical Memories \citep{pouransari2026pretraining} & Language modeling & Pre-deploy & Persist. & \cellH & \cellL \\
    Memory Layers at Scale \citep{berges2024memory} & Language modeling & Pre-deploy & Persist. & \cellH & \cellL \\
    PlugLM \citep{cheng2023decouple} & Knowledge injection & Pre-deploy & Persist. & \cellH & \cellL \\
    RING \citep{xu2026ring} & Knowledge injection & Pre-deploy & Persist. & \cellH & \cellL \\
    PKM \citep{lample2019large} & Language modeling & Pre-deploy & Persist. & \cellH & \cellL \\
    \midrule
    \multicolumn{6}{@{}l}{\textbf{Hybrid --- Online}} \\
    \addlinespace[1pt]
    SHINE \citep{liu2026shine} & Task adaptation & Per-request & Ephem. & \cellM & \cellL \\
    Temp-LoRA \citep{wang2024greater} & Long-context modeling & Per-request & Ephem. & \cellM & \cellL \\
    Absorber LLM \citep{zhang2026absorber} & Test-time training & Per-token & Session & \cellL & \cellL \\
    Doc2Atom \citep{diao2026doc} & Document QA & Per-request & Ephem. & \cellM & \cellL \\
    Code2LoRA \citep{hotsko2026code2lora} & Task adaptation & Per-request & Ephem. & \cellM & \cellL \\
    \addlinespace[2pt]
    \multicolumn{6}{@{}l}{\textbf{Hybrid --- Offline}} \\
    \addlinespace[1pt]
    LoRA \citep{hu2021lora} & Task adaptation & Pre-deploy & Persist. & \cellH & \cellL \\
    Macaron-V1 \citep{mindlab2026macaronv1} & Task adaptation & Pre-deploy & Persist. & \cellH & \cellL \\
    LatentSkill \citep{yu2026latentskill} & Task adaptation & Pre-deploy & Persist. & \cellH & \cellL \\
    KnowLa \citep{luo2024knowla} & Knowledge injection & Pre-deploy & Persist. & \cellH & \cellL \\
    LAG \citep{fleshman2025lora} & Knowledge injection & Pre-deploy & Persist. & \cellH & \cellL \\
    GeoRA \citep{zhang2026geora} & Task adaptation & Pre-deploy & Persist. & \cellH & \cellL \\
    \bottomrule
    \end{tabularx}
    \end{table}

\end{document}